\documentclass{article}

\usepackage[english]{babel}

\usepackage[letterpaper,top=2cm,bottom=2cm,left=3cm,right=3cm,marginparwidth=1.75cm]{geometry}

\usepackage{amsmath}
\usepackage{graphicx}
\usepackage[colorlinks=true, allcolors=blue]{hyperref}
\date{v3: 07.03.2024}

\title{Embracing Large Language and Multimodal Models for Prosthetic Technologies}
\author{Sharmita Dey\thanks{corresponding author, conceptualized and led the ideas presented\\
sharmita.dey@cs.uni-goettingen.de},
Arndt F. Schilling}

\begin{document}
\maketitle

\begin{abstract}
This article presents a vision for the future of prosthetic devices, leveraging the advancements in large language models (LLMs) and Large Multimodal Models (LMMs) to revolutionize the interaction between humans and assistive technologies. Unlike traditional prostheses, which rely on limited and predefined commands, this approach aims to develop intelligent prostheses that understand and respond to users’ needs through natural language and multimodal inputs. The realization of this vision involves developing a control system capable of understanding and translating a wide array of natural language and multimodal inputs into actionable commands for prosthetic devices. This includes the creation of models that can extract and interpret features from both textual and multimodal data, ensuring devices not only follow user commands but also respond intelligently to the environment and user intent, thus marking a significant leap forward in assistive technology. 


\end{abstract}

\section{Preface}

This article diverges from the conventional format of academic papers, embodying instead a position article that lays bare our vision and aspirations for the future of prosthetic devices.  It is a continually developing article, reflecting an evolving vision for a prosthetic device that seamlessly integrates the capabilities of large language models and multimodal inputs. This ambition is not just to create a tool that compensates for physical limitations but to facilitate an intelligent complement that understands and anticipates the user’s interoceptive and exteroceptive needs, thereby enhancing human-machine synergy. Through this exploration, we seek to initiate a dialogue within the scientific community and beyond, challenging us to rethink the boundaries of assistive technology and its potential to enrich the human experience.

\section{Introduction}

The advent of large language models (LLMs) like the text-based DaVinci or the GPT series, as well as Large Multimodal Models (LMMs), has marked a significant paradigm shift in the field of natural language processing and beyond. LLMs, trained on vast datasets, have shown extraordinary proficiency in understanding and generating human-like text, revolutionizing our interaction with technology through natural language prompts. LMMs, on the other hand, extend these capabilities by integrating and interpreting multiple data types, such as images, videos, and audio, alongside text. This integration allows for a more comprehensive understanding and response to complex inputs.
Our aim is to leverage these breakthroughs for a noble and innovative purpose: to control assistive prosthetics through the power of natural language and multimodal commands. To our knowledge, as of now, there has been no introduction of either LLMs or LMMs in the realm of prosthetic technology. This unexplored territory opens up immense possibilities for enhancing the functionality and user-friendliness of prosthetic devices.
Currently, intelligent assistive devices such as prostheses or orthoses primarily operate on predefined commands encoded as state machines or, less commonly, through learned models that interpret different states or modes of gait from sensor commands like proprioception or electromyography (EMG). In such scenarios, user intervention or customization is not common. Moreover, existing human-in-the-loop approaches primarily focus on model-based refinements, where the model adjusts itself based on online gait parameters or walking trials from humans. The concept of adjusting an assistive device fluidly using natural language commands remains unexplored. Both LLMs and LMMs, present an opportunity to revolutionize this approach by enabling more dynamic and user-responsive control mechanisms.

Traditional methods like switches and button controls, though user-directed, offer limited interaction, typically requiring technical knowledge or memorization of parameters. These methods essentially mirror the limitations of finite-state machines. In contrast, imagine the ease and efficiency if a user could instruct their device in natural language, saying something like, "I want to walk on a rocky terrain with occasional wet surfaces, adjust the device properties accordingly for future steps." Such an approach would seamlessly integrate sophisticated model-based control, enhancing the user experience significantly. An LMM could further enhance this interaction by analyzing additional inputs, like images or sounds of the terrain, and EMG signals during user activity to make more informed adjustments. This would integrate sophisticated model-based control with an intuitive user interface, significantly enhancing the user experience.
To encapsulate, the introduction of LLMs and LMMs in this domain could radically transform how users interact with their prosthetic devices. By allowing users to issue natural language commands, these devices could be made more intuitive and user-friendly. This application of LLMs and LMMs in prosthetic technology is a prime example of how AI can be used to enhance human capabilities and improve the quality of life for individuals with disabilities. It represents a shift from the user needing to understand and adapt to the technology, to the technology being able to understand and adapt to the user's needs. This could lead to more personalized, adaptable, and intuitive assistive devices, ultimately making them more effective and easier to use.

\section{Core Concept}

Our core innovation lies in utilizing LLMs to interpret and execute gait tasks and control commands for prosthetic technologies. This method allows users to interact with their devices using natural language commands, thereby removing the need for an in-depth understanding of the technical specifications. The system aims to process a broad spectrum of verbal and written instructions, converting them into precise control signals for the prosthesis, orthosis, or exoskeleton. This includes interpreting commands for specific movements, adjusting gait patterns, or responding to changing environmental conditions.
Enhancing this concept further, Large Multimodal Models (LMMs) akin to models like Gemini could significantly enrich the process of controlling assistive devices. LMMs have the capability to process and integrate information from multiple sources, not just text. By combining natural language commands with other forms of data, such as visual inputs from an onboard camera on the assistive device, or auditory information depicting the sounds of different terrains, EMG signals from the user's muscle activations, these models can gain a more comprehensive understanding of the user's environment and locomotive intent.
For example, if a user is about to walk on a sandy beach, they could issue a verbal command while the device's camera captures images of the sand. The LMM could then process both the spoken command and the visual data to make more informed adjustments to the device. It could modify various parameters like the torque, stiffness, damping, or grip of a prosthetic foot to ensure optimal performance on the sandy terrain.
This multimodal approach would enable the device to not only understand the user's instructions but also to perceive the locomotive intent and environment in a way similar to human senses, leading to a more detailed and responsive adaptation. Such an integration of LMMs in assistive technology could dramatically improve the adaptability and effectiveness of these devices, offering users a more natural and seamless experience in navigating a variety of terrains and environments.

\section{Realization of the Idea}
The realization of this approach to control assistive devices through large language and multimodal models involves several key stages and technical strategies. Here, we delve deeper into the specifics of each implementation method.

\subsection{LLM-based Control Model}

The idea is to utilize a pre-trained LLM to extract encoder features for new text corpora. These features can be mapped to corresponding gait commands, like impedance, stiffness, torque, and velocity profiles by training a decoder over the encoder features. 

\begin{enumerate}
    \item Data preparation: The first step involves gathering a diverse and comprehensive dataset that includes a range of natural language commands to be mapped to gait and movement commands.
    \item Encoder feature extraction: We can utilize a pre-trained LLM encoder to extract relevant features from the natural language input. These features represent the essence of the user's command in a format that can be used for further training for the control system of the assistive device. 
    \item Decoder training: A crucial component of this system is the decoder layer, which translates the feature set from the LLM into actionable control signals for the device. This decoder must be robust and adaptable, able to accurately interpret a wide range of commands and convert them into precise mechanical adjustments. The extracted features are mapped to specific control commands by training the decoder, such as adjustments in the device's impedance, stiffness, torque, or velocity. The decoder could be realized using a lightweight transformer model that attends to the encoder memory and the decoder inputs to learn a mapping from these encoder features to the specific gait commands for adjusting the assistive/rehabilitation device.
\end{enumerate}

\subsection{LMM-based Control Model}

Utilize a pre-trained LMM to extract encoder features for new multimodal data. Using multimodal inputs like natural language commands and corresponding images, map to the same physical parameters, applying similar principles as above.

\begin{enumerate}
    \item Integration of multimodal inputs: For LMM-based models, the process begins with the integration of multiple input modalities. Alongside textual commands, this includes other exteroceptive inputs (such as images or videos of the terrain) and/or interoceptive inputs such as proprioception, and EMG, and other relevant sensory data.

    \item Encoder feature extraction: We can either utilize a pre-trained LMM encoder to extract relevant features from the multimodal input and further augment it with additional modalities specific to prosthesis control e.g., EMG, or train our own encoder from scratch. The encoded features represent a condensed aggregation of the user's command and the inputs from the other modalities in a format that can be used for further training for the control system of the assistive device. 
    The multimodal backbone can be a standard transformer-based model to understand the interactions between the different modalities and produce the required representation to be decoded.

    \item Decoder training: Similar to the LLM-based decoder training, the extracted multimodal features are mapped to specific control commands by training the decoder, such as adjustments in the device's impedance, stiffness, torque, or velocity.
\end{enumerate}

\begin{figure}
\centering
\includegraphics[width=0.8\linewidth]{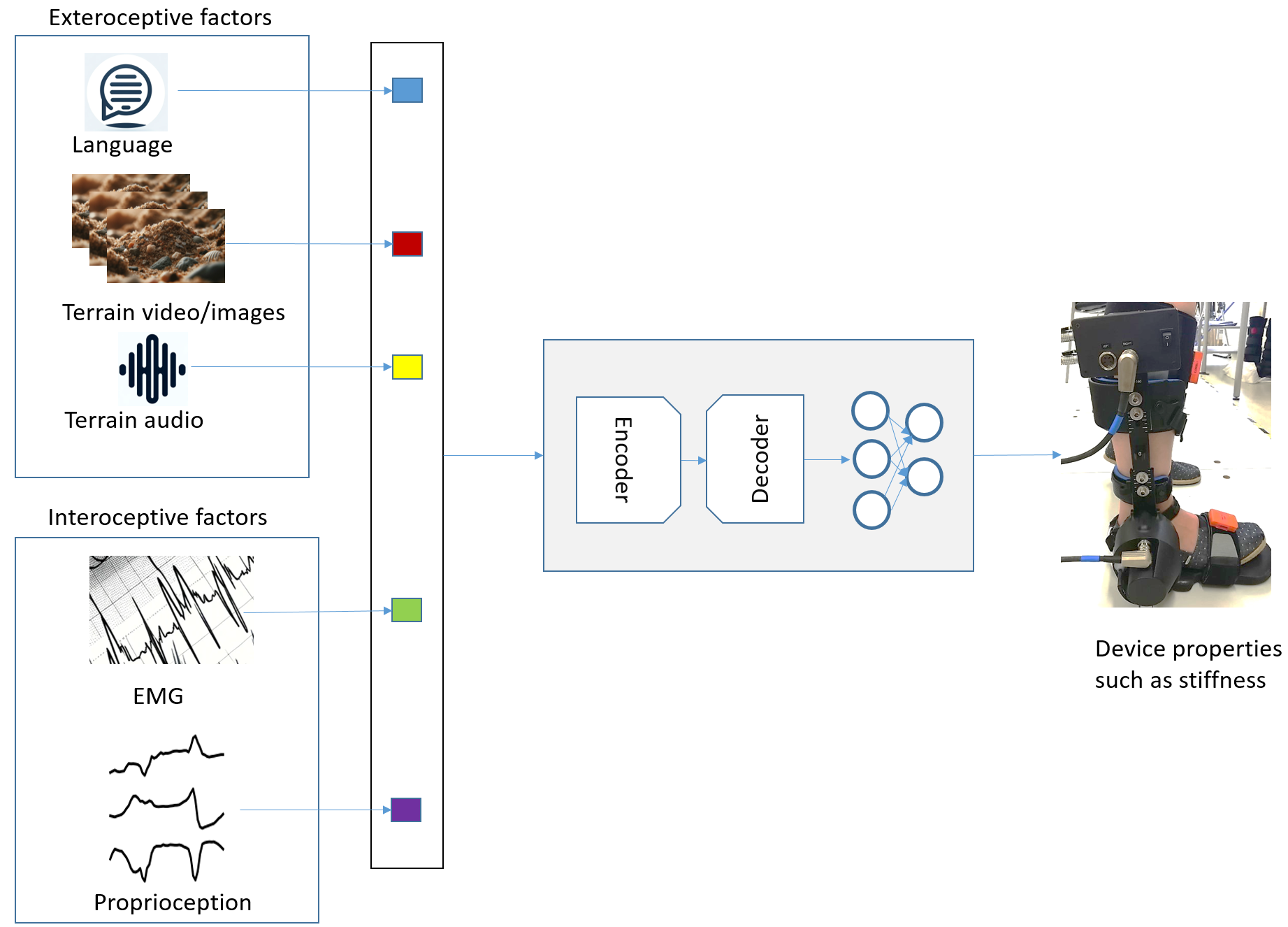}
\caption{Overview of the LMM-based model for exoskeleton control. The different color tokens represent different encoded input modalities from various sources. }
\end{figure}

\section{Adapting LLMs and LMMs for Practical Embodiment in Rehabilitation Technologies}

\subsubsection{Background} LLMs have showcased huge proficiency in understanding and responding to natural language instructions to formulate further plans \cite{huang2022inner,ahn2022can}. However, as \cite{ahn2022can} noted, although LLMs are capable of producing high-level semantically meaningful text responses to a certain scenario, it does not often translate to real-world executable actions for an agent to perform. To effectively tailor language models for such use cases, we need to guide them towards converting high-level instructions into a sequence of executable low-level skills. Although careful prompt engineering could help steer a language model towards a desired response format, this involves crafting examples within the prompt text that clearly define the task and the expected structure of the model's response, and moreover, this still does not ensure a suitable actionable command to be executed \cite{ahn2022can}. 
\subsubsection{Previous Developments} In this context, the authors of \cite{ahn2022can} capitalize on the ability of the LLM not just to interpret instructions, but also to evaluate the likelihood that the LLM generated instruction will contribute effectively towards the completion of the given high-level instruction using skill scoring and affordance functions that quantify the likelihood of the skill's success from the current state, potentially using a learned value function. The value provided by this affordance function is then used to weight the likelihood score assigned to each skill by the LLM. Essentially, the LLM assesses the probability that a particular skill will help in fulfilling the instruction, while the affordance function estimates the probability of that skill's successful execution given the current conditions.
\subsubsection{Wearable Rehabilitation Device Control} In the realm of controlling prosthetics, orthoses, and exoskeletons, we propose direct interpretation and supervision of language and multimodal inputs, including EMG, inertial measurement units, video, and audio, to formulate executable real-world commands for these assistive devices. Our methodology aims to build upon the foundational strengths of pre-trained LLMs and LMMs.  We plan to harness the power of embeddings generated from these models, converting them into actionable commands tailored for assistive prosthetics.  Furthermore, our utilization of multimodal models aims to decrease non-conformity between expected and desired behavior which is also unexplored in this field.
By integrating these advanced artificially intelligent models, the system can learn and adapt to the user's unique preferences and requirements over time, ensuring a continually improving and personalized experience. This adaptability not only enhances the functionality of the assistive prosthesis but also fosters a deeper bond between the user and the technology, as the system evolves to become more attuned to the user's needs and lifestyle. This approach stands to revolutionize the way we think about and interact with assistive prosthetics, opening new horizons for accessibility and user empowerment. Recent \cite{brohan2023rt} and ongoing efforts \cite{octo_2023} on creating generalist robot policies from multimodal inputs such as language, images, and proprioception resonate with our vision. Similar open-source models \cite{octo_2023} can be leveraged as a pre-trained base to further develop our models or augment them with additional modalities, e.g., EMG.

\bibliographystyle{ieeetr}
\bibliography{sample}

\end{document}